\documentclass[10pt,twocolumn]{article}

\usepackage{iccv}
\usepackage{times}
\usepackage{epsfig}
\usepackage{graphicx}
\usepackage{amsmath}
\usepackage{amssymb}
\usepackage{subfigure} 
\usepackage{color}

\usepackage[pagebackref=true,breaklinks=true,colorlinks,bookmarks=false]{hyperref}

\iccvfinalcopy 

\def\iccvPaperID{2097} 

\ificcvfinal\fi

\begin{document}

\title{Anisotropic EM Segmentation by 3D Affinity Learning and Agglomeration}
\author{Toufiq Parag$^1$, Fabian Tschopp$^4$, William Grisaitis$^2$, Srinivas C Turaga$^2$, Xuewen Zhang$^5$ \\Brian Matejek$^{1}$,  Lee Kamentsky$^1$, Jeff W. Lichtman$^3$, Hanspeter Pfister$^1$\\
 \\
       $^1$School of Engg and Applied Sciences, Harvard University, Cambridge, MA\\
       $^2$Janelia Research Campus, Ashburn, VA\\
       $^3$Dept of Molecular and Cellular Biology, Harvard University, Cambridge, MA \\
       $^4$ Institute of Neuroinformatics, University of Zurich and ETH Zurich, Switzerland.\\
       $^5$ Chester F. Carlson Center for Imaging Science, RIT, Rochester, NY.
       \\
        \\
       email: paragt@seas.harvard.edu\\ 
}


\maketitle
\thispagestyle{empty}


\begin{abstract}

The field of connectomics has recently produced neuron wiring diagrams from relatively large brain regions from multiple animals. Most of these neural reconstructions were computed from isotropic (e.g., FIBSEM) or near isotropic (e.g., SBEM) data. In spite of the remarkable progress on algorithms in recent years, automatic dense reconstruction from anisotropic data remains a challenge for the connectomics community. One significant hurdle in the segmentation of anisotropic data is the difficulty in generating a suitable initial over-segmentation. In this study, we present a segmentation method for anisotropic EM data that agglomerates a 3D over-segmentation computed from the 3D affinity prediction. A 3D U-net is trained to predict 3D affinities by the MALIS approach. Experiments on multiple datasets demonstrates the strength and robustness of the proposed method for anisotropic EM segmentation.

\end{abstract}

\section{Introduction}

In past few years, connectomics has grown to become a mature field of study in neuroscience. Reconstruction of neural circuits from Electron Microscopic (EM) images of animal brain is not a hypothetical concept anymore -- multiple attempts in this field have already furnished the neuroscience community with wiring diagrams from different animals~\cite{takemura13, helmstaedter13a, kim2014nature, takemura15pnas, takemura17elife}. These studies, and others e.g.,~\cite{takemura17elife2}, report crucial biological discoveries stemming from the computed wiring diagrams. While electron microscopy is capable of providing the most exhaustive knowledge about the cellular anatomy and connectivity among all other imaging techniques, it also produces an enormous amount of data that is too large to process manually. All the aforementioned works adopt a semi-automated strategy where the results of automated algorithms are manually corrected afterwards.

Extraction of neural shapes entails a 3D segmentation of EM data volume, i.e., tracing of cellular processes within and across different sections/planes of the EM volume. With the same resolution in all x, y, z dimensions, isotropic images can capture the continuity in z dimension (or in depth) more than other imaging approaches. In an isotropic EM volume, a cellular process almost never overlaps (to a significant extent) with that from another neuron across different sections. This characteristic of isotropic recording offers a fundamental advantage for the automatic 3D segmentation methods. Another benefit of the isotropic, or near isotropic, imaging techniques such as FIBSEM~\cite{takemura15pnas, takemura17elife} and SBEM~\cite{helmstaedter13a,kim2014nature} is that they typically give rise to little or no staining and imaging artifacts. 

Not surprisingly, most of the successful efforts for neural reconstruction were performed on isotropic or near isotropic data. SBEM has a voxel resolution of $16 \times 16 \times 25$nm in x, y, z respectively which we consider to be very close to being isotropic for practical reconstruction purposes. The success of these efforts can be largely contributed to the progress in the 3D segmentation algorithms that have been developed recently~\cite{turaga09malis, turaga2010affinity, jain11, iglesias13, parag15b, parag14, andres08, andres12}. In addition to the improved methods,  the profound improvement in automatic processing accuracy in the studies of \cite{takemura15pnas, takemura17elife}, that led to more than 5 times speed up in overall reconstruction time compared to ~\cite{takemura13}, can also be partially attributed to the transition to isotropic FIBSEM images. However, isotropic imaging has its limitations. FIBSEM, for example, is not ideal for large scale imaging in the range of several hundreds cubic microns~\cite{hayworth12ijmc} . There have been experiments~\cite{hayworth12hotknife} for scaling up the volume that FIBSEM can capture successfully, but have not yet delivered a large scale connectome. 

On the other hand, anisotropic approaches such as MSEM~\cite{eberle15multibeam} are capable of imaging volume in cubic millimeter scale and therefore is suitable for large scale connectomics. Anisotropic imaging records images from tissue sections thicker than those isotropic methods (e.g., FIBSEM) can capture. While this strategy enlarges the brain region that can be captured by the same volume of data, it decreases the continuity in z dimension significantly. As a result, the area that pertains to one cellular process on any particular plane can overlap with multiple processes across different planes. This gives an additional challenge to the segmentation process -- without an additional mechanism specifically designed to handle such situation, applying 3D segmentation in a straightforward fashion will inaccurately merge many neurons into one large body. 

Perhaps a natural idea to operate on this data is to apply 2D segmentation on each section~\cite{ciresan12, jurrus10mia, iglesias13, seymour14rhoana, meirovitch2016multi, Matveev2017Multicore} and then link the 2D segments on different sections with a linkage or cosegmentation algorithm~\cite{vazquez11,funke12,beier2016}. However, as the results from the EM segmentation challenges SNEMI (\url{brainiac2.mit.edu/SNEMI3D}), CREMI (\url{cremi.org}) and many experiments across different research groups suggest, such an approach typically lead to a level of over-segmentation that is not very favorable for efficient reconstruction. Our observation indicates that if the input over-segmentation is largely fragmented or under-segmented, one cannot expect a high quality solution from these algorithms despite their solid conceptual and theoretical foundation. It appears to us that generating the initial over-segmentation remains a challenge for the anisotropic EM volumes.  

In this paper, we present a method to compute a 3D over-segmentation of anisotropic data by  learning 3D affinities directly using a deep neural network. Given such an input over-segmentation we apply the agglomeration method of~\cite{parag15b} to generate the final segmentation results. We train a 3D U-net~\cite{ronneberger15unet} using MALIS~\cite{turaga09malis} loss to predict the 3D affinities in $x, y, z$ dimensions. This particular loss function emphasizes on learning the sparse locations that are more important than other to preserve neuron topology rather than imposing equal weight on all pixels. In addition to~\cite{turaga09malis}, multiple other studies~\cite{parag2015ICCV,jain10cvpr} demonstrated the importance on training at sparse topologically important locations for EM connectomics. In addition, the experiments in ~\cite{parag15b} and~\cite{seymour14rhoana}  suggests that agglomerative clustering can achieve equivalent or better segmentation accuracy than linkage algorithms~\cite{vazquez11,andres12} given a 3D over-segmentation with minimal or no false merges. Extensive experiments on multiple anisotropic dataset exhibit superior performance of the 3D affinity learning and agglomeration compared to standard methods. The proposed method produces impressive improvement over the existing approaches both qualitatively and quantitatively as we report them in the result section. Another study by Funke et.al. also reports similar findings independently on different datasets. This approach uses a different agglomeration technique as well, however the quality of their segmentation is also very impressive. 

\section{3D Affinity Learning and Over-segmentation}

Given an anisotropic volume, we learn and predict the 3D affinities among the voxels within the volume. In particular, instead of identifying whether or not one particular voxel belong to cell boundary (or membrane),  we decide whether or not any pair of voxels $\{~\{x, y, z\}, \{x+1, y, z\}~\}$ -- or  $\{~\{x, y, z\}, \{x, y+1, z\}~\}$ and $\{~\{x, y, z\}, \{x, y, z+1\}~\}$ -- reside within a cell and therefore needs to be connected together to form the segments. Learning affinities has been popularized in connectomics by the past works of~\cite{turaga09malis, Huang14,lee15nips}. We apply the MALIS training method proposed in ~\cite{turaga09malis} in this work.

The MALIS learning algorithm emphasizes on learning the affinities that are critical to preserve the neural anatomy. Instead of training on affinities between voxels from all possible pairs in a volume, it locates the the edge that either incorrectly splits the path between two voxels from one cell or incorrectly merges voxels from different cells. These edges are often called the maximin edges (conversely, minimax or  bottleneck edges of they are defined on costs) and can be efficiently computed for all possible paths between two voxels within a volume from a minimal spanning tree (MST)~\cite{hu61or, camerini78mst, turaga09malis}. Topologically, these edges are more important than others for correct segmentation of neural shapes and therefore must be emphasized in the learning algorithm. In connectomics, multiple other studies have also discovered this phenomenon and trained a classifier with more (sometimes exclusive) attention to sparse but topologically important samples~\cite{jain10cvpr,parag2015ICCV}. Although the mechanism by which these samples are selected and learned is different in each of these works.

We train a 3D U-net~\cite{ronneberger15unet} for learning the affinities with MALIS loss. The particular architecture we utilize extends the Caffe model for 3D convolutions and bypass connections in U-nete and is publicly available at \url{https://github.com/naibaf7/PyGreentea}. One notable aspect of the output ( and target label) of the U-net is we only learn the affinities of 3 consecutive sections. That is, the z-affinities we compute only connects one voxel to the adjacent voxels in the preceding and following sections only. In our experiments, training z-affinities only for the two neighboring sections resulted in the similar segmentation results as those produced by long range (>3 sections) z-affinities.

Given the $x, y, z$ affinities produced by the U-net, we apply the Z-watershed algorithm~\cite{zlateskiS15} with suitable size parameters for different datasets.

\section{3D agglomeration}

The 3D over-segmentation generated by Z-watershed is then refined by an agglomeration algorithm presented in~\cite{parag15b}. Given an over-segmented volume with minimal or no false merges in it, an agglomeration algorithm repeatedly merges two neighboring supervoxels or fragmented regions into one.  A supervoxel boundary classifier decides which supervoxels should be merged to reduce the fragmentation as well as which ones should be left separated for accurate identification of neuron structure. The classifier is trained to make a decision based on features computed on each of the supervoxels and the boundary separating them. The process is hierarchical, that is, after each merge operation, the boundaries and the features  are recomputed for the newly created supervoxel. The study of~\cite{parag15b} reported experimental evidence that the hierarchical agglomeration leads to better segmentation accuracy that the optimization based non hierarchical method of~\cite{andres12}. Another paper~\cite{seymour14rhoana} also suggested using agglomerative clustering of~\cite{iglesias13} produced improved segmentation results compared to fusion method~\cite{vazquez11}.

\section{Experiments and Results}
We have applied our method multiple datasets. Our findings from these experiments are summarized in the following sections. In all of our experiments, we used a 3D U-net with depth 3, very similar to the author's architecture~\cite{ronneberger15unet}. The most important difference between our network and the author's version is we only pool in x and y dimensions only, and not in z, in each downsampling layer. The input and output to the network are volumes of grayscale EM stacks of sizes $[204, 204, 33]$ and  $[116, 116, 3]$ respectively. Network trained with more z sections (input 44 and output 16) did not increase the accuracy. From different iterations of the MALIS training, we pick the iteration at which the network produces the least under-segmented output. For supervoxel training, the features were computed on all 3 affinities predicted in the x, y, z dimensions and we used the features, training strategy of~\cite{parag15b} (without mitochondria information).

\subsection{Mouse Neocortex data~\cite{kasthuri15cell}}
We have tested this method on the MSEM (single beam microscope) images collected from mouse cerebral cortex as discussed in~\cite{kasthuri15cell}. The pixel resolution for these dataset was $3 \times 3 \times 30$nm in x, y, z respectively. However, for our experiments, the images was downsampled by a factor of 2 in x, y dimensions. For training, a volume of $1024 \times 1024 \times 100$ pixels (approx $6 \times 6 \times 3 ~\mu^3$ ) was cropped from the manually annotated parts of~\cite{kasthuri15cell} data. Both the affinity predictor 3D U-net and the supervoxel boundary classifier were trained on this volume. 

\begin{figure}[h]
\vspace{-0.1cm}
\begin{center}
\includegraphics[width=0.84\columnwidth, height=0.75\columnwidth]{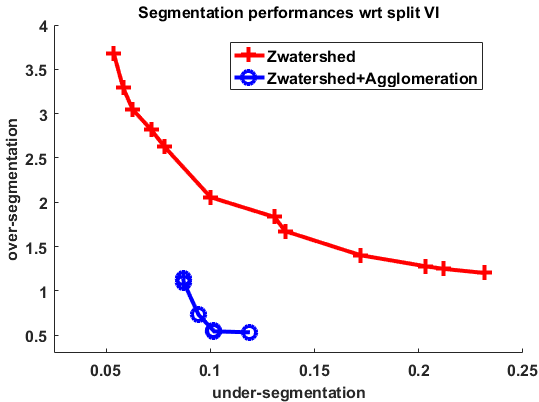}
\caption{\scriptsize Quantitative evaluation of competing methods on the $ 6 \times 6 \times 4~\mu^3 $ test volume. The x and y axes correspond to the under- and over-segmentation respectively. Red - only Z-watershed~\cite{zlateskiS15} on 3D affinities, and, blue - Z-watershed + 3D agglomeration~\cite{parag15b}. Lower curves correspond to better result than the those corresponding to the curves above.}\label{F:KASTHURI_WSHED}
\end{center}
\vspace{-0.7cm}
\end{figure}

In the first experiment, we examine the quality of the segmentation yielded by only the 3D affinity prediction and the Z-watershed~\cite{zlateskiS15} method to understand whether or not a subsequent agglomeration is necessary. A volume of $1024 \times 1024 \times 150$ pixels (approx $6 \times 6 \times 4 ~\mu^3$ ) was utilized for this test. The segmentation performance is evaluated using the split-VI mesaures, i.e., the two quantities of the Variation of Information (VI) metric that correspond to uner- and over-segmentation errors. Following~\cite{iglesias13,parag15b, parag2015ICCV}, the VI quantities pertaining to under and over-segmentation are plotted on x and y axes respectively. An ideal result should have a $(0, 0)$ error and should be placed at the origin of this plot. Please note that the \emph{under-segmentation error is more pronounced in the spli-VI plots since false merges are far more detrimental to the overall neural reconstruction accuracy than over-segmentation}. In Figure~\ref{F:KASTHURI_WSHED}, the segmentation errors from only Z-watershed output  and that of Z-watershed and agglomeration at different thresholds are plotted with red and blue respectively. The split-VI values suggests that one could achieve significantly better segmentation by applying 3D agglomeration on the result of Z-watershed.  

\begin{figure*}[t]
\vspace{-0.1cm}
\begin{center}
\subfigure[Split-VI difference with baseline]{\includegraphics[width=0.85\columnwidth, height=0.75\columnwidth]{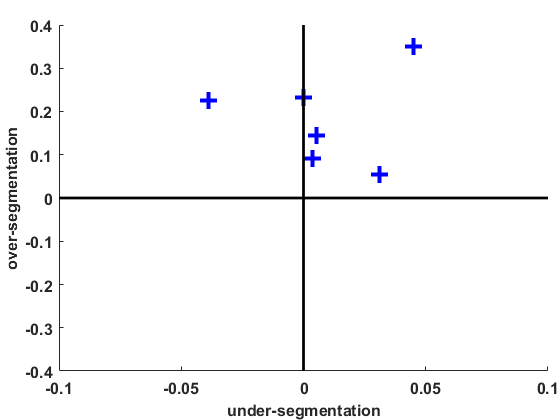}\label{F:KASTHURI_COMPARISON_RHOANA}}
\subfigure[Split-VI difference with VD2D3D]{\includegraphics[width=0.85\columnwidth, height=0.75\columnwidth]{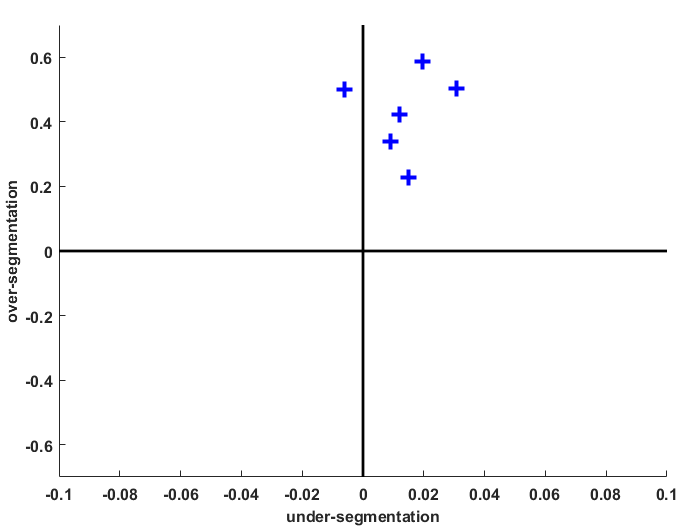}\label{F:KASTHURI_COMPARISON_2D3D}}
\caption{\scriptsize The split-VI difference between the proposed method and baseline algorithms based on Rhoana on 6 volumes. Values on the first quadrant indicates the output of the proposed methods is better both in terms of under- and over-segemtnation.}\label{F:KASTHURI_COMPARISON}
\end{center}
\vspace{-0.7cm}
\end{figure*}

The study of~\cite{kasthuri15cell} annotated  a relatively large volume for their biological analyses. For a more comprehensive test, we collected 4 volumes of size $6 \times 12 \times 3 \mu^3 $ and 2 volumes of size $ 6 \times 12 \times 2.4 \mu^3$ manually labeled volumes and compared our method with a baseline algorithm similar to that presented in the Rhoana pipeline paper~\cite{seymour14rhoana}. For the baseline method, a 2D U-net was trained to predict the pixel membrane probabilities. These 2D probabilities were then used to generate a 3D over-segmentation by waterhsed on dilated membrane predictions. The parameters for the dilation and watershed were tuned to minimize the false merges. This over-segmentation is then  refined by 3D agglomeration of~\cite{parag15b}. The paper~\cite{seymour14rhoana} reported improved result with agglomeration over fusion based method~\cite{vazquez11} on anisotropic data.

For comparison, we plot the differences in split-VI~\cite{parag15b,  iglesias13} between the baseline and the the proposed method, $\text{VI}_{\text{baseline}} - \text{VI}_{\text{proposed}}$ in Figure~\ref{F:KASTHURI_COMPARISON_RHOANA}. Because it will be difficult to compare the curves at different agglomeration thresholds, we selected the parameters with the lowest error for both methods. Also, the plot is stretched more in the x-axis than y to emphasize the under-segmentation error more than the false splits.  If the difference lies in the first quadrant (clockwise), the result from proposed method is more accurate than those of baseline method in terms of both over- and under-segmentation.

In addition, we also compared our results on these six volumes with those produced by the VD2D3D affinity prediction~\cite{lee15nips} followed by Z-watershed and agglomeration. The VD2D3D network and the supervoxel classifier were trained on the same $6 \times 6 \times 3~\mu^3$ volume that the proposed method uses. We plot the split-VI difference with VD2D3D prediction and agglomeration in Figure~\ref{F:KASTHURI_COMPARISON_2D3D}. As the plot suggests, the proposed method achieves superior accuracy to both baseline and VD2D3D in both under- and over-segmentation in almost all the volumes.   

\begin{figure}[h]
\vspace{-0.1cm}
\begin{center}
\includegraphics[width=0.42\columnwidth, height=0.75\columnwidth]{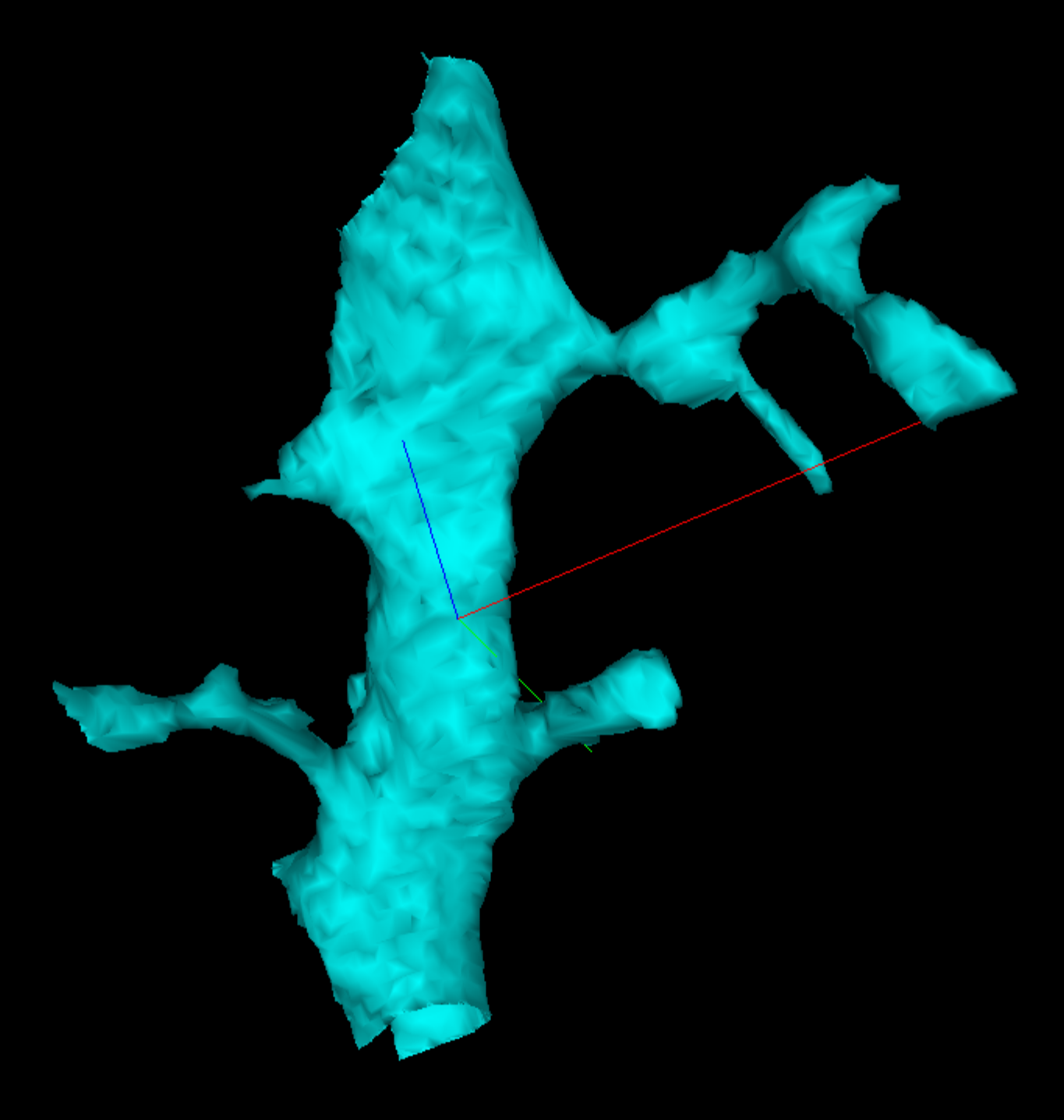}
\includegraphics[width=0.42\columnwidth, height=0.75\columnwidth]{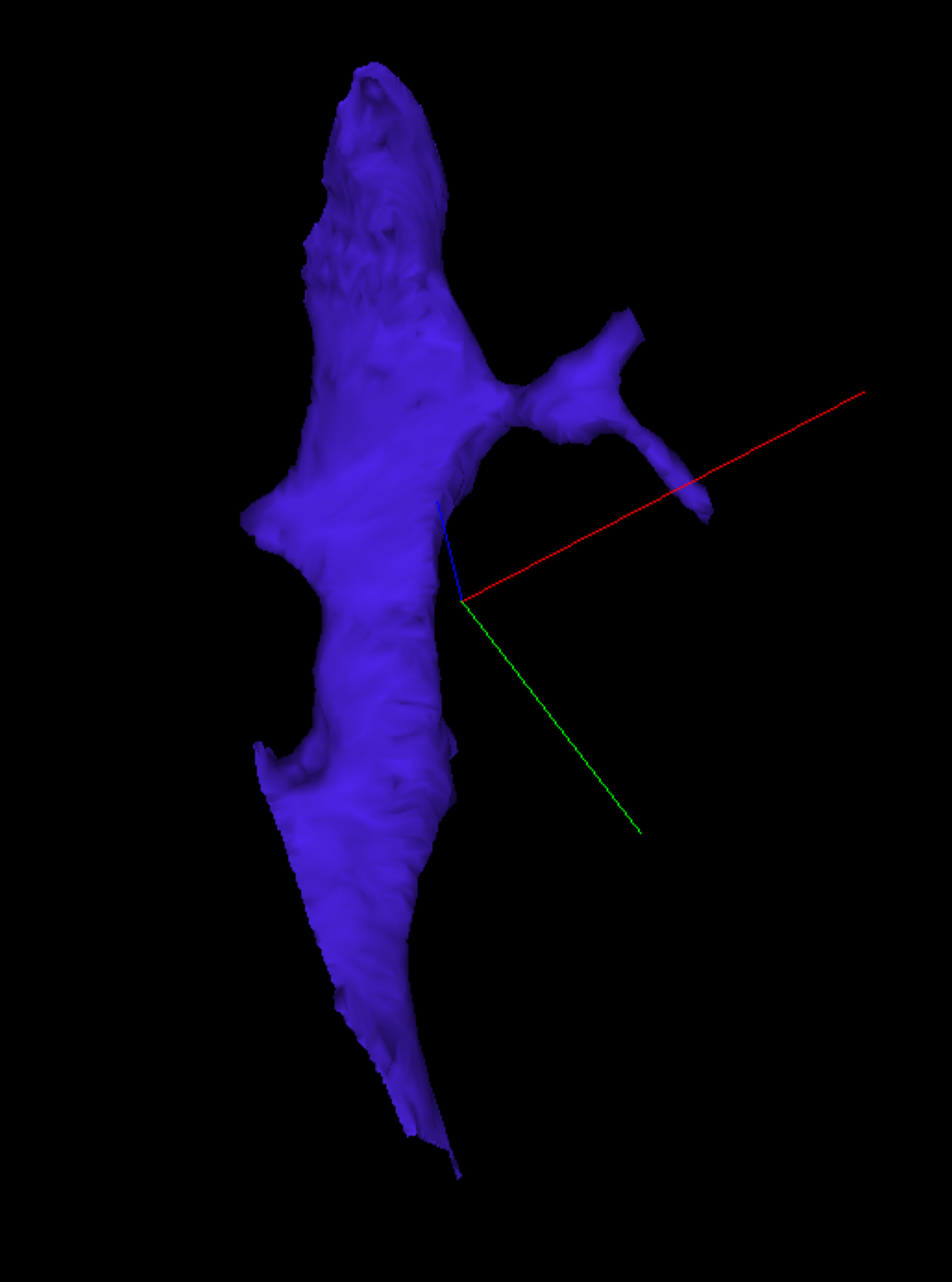}
\includegraphics[width=0.42\columnwidth, height=0.75\columnwidth]{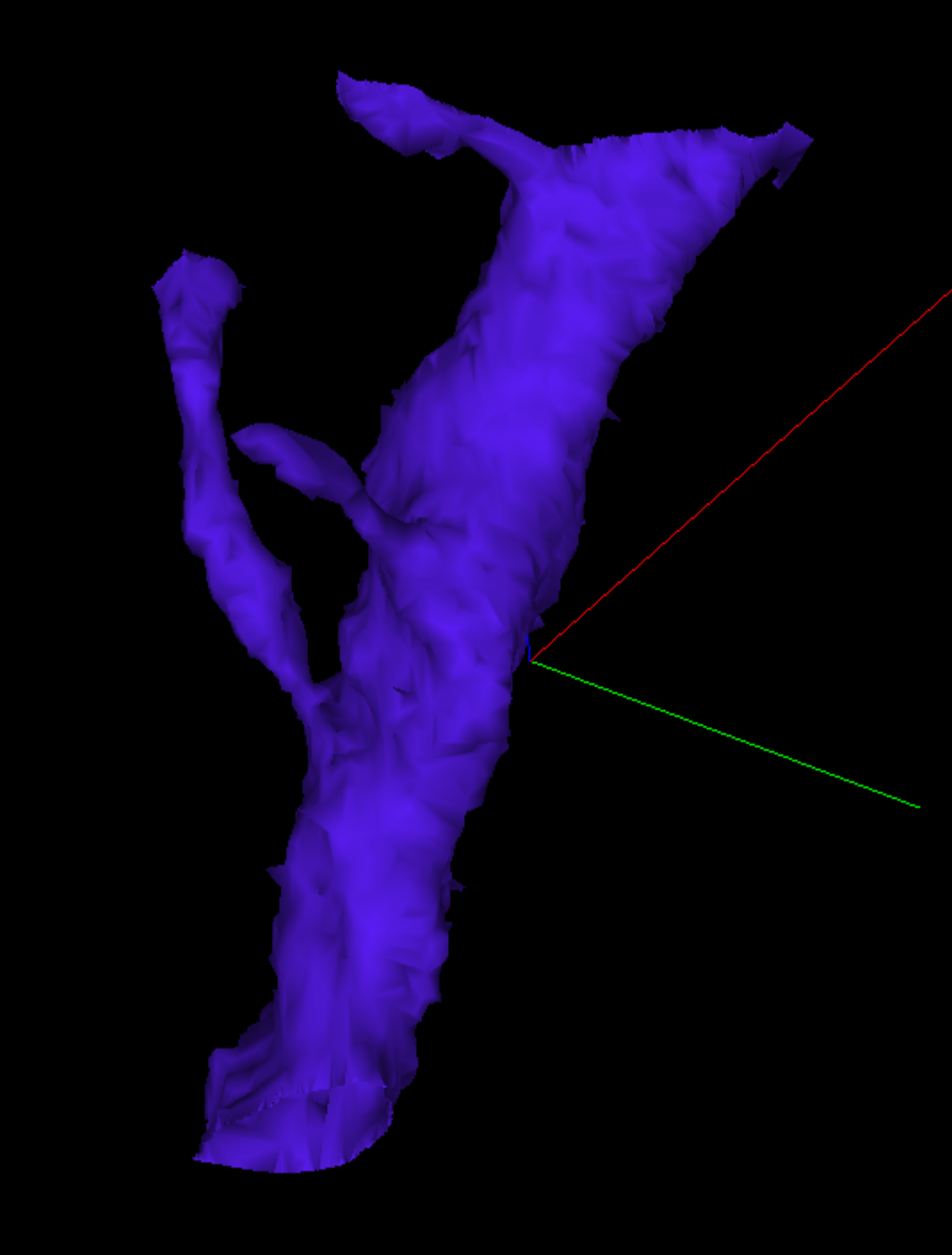}
\includegraphics[width=0.42\columnwidth, height=0.75\columnwidth]{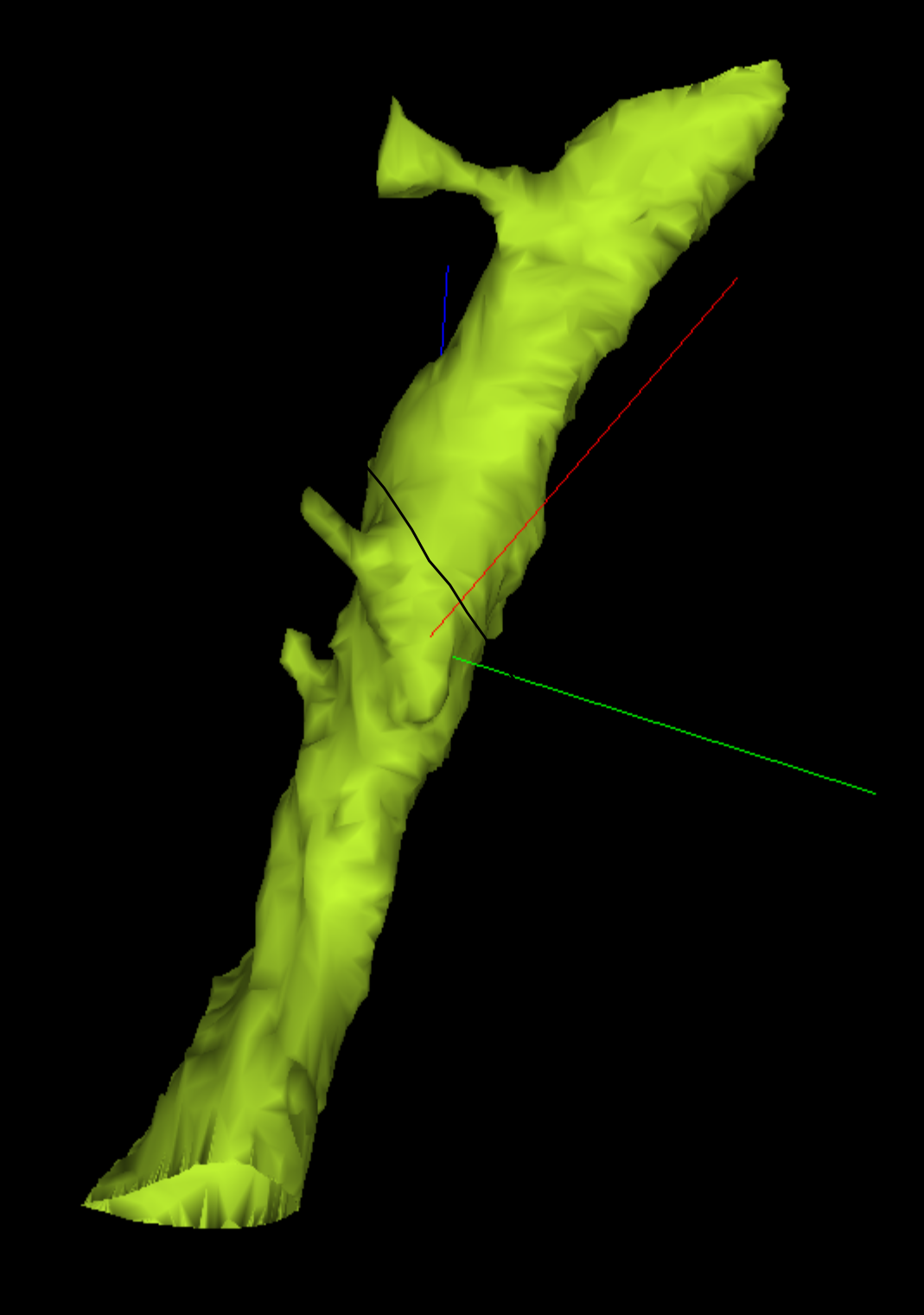}
\caption{\scriptsize 3D views of the segmentation results from the proposed method (left column) and the baseline (Rhoana based) technique (right column). The proposed approach can trace the thinner processes, e.g., spine necks, more accurately than the baseline method. }\label{F:KASTHURI_QUALITATIVE}
\end{center}
\vspace{-0.7cm}
\end{figure}

Figure~\ref{F:KASTHURI_QUALITATIVE} compares 3D views of neural reconstructions from the proposed and the baseline algorithms (based on Rhoana) on one of the $6 \times 12 \times 3 \mu^3$. On each row, the left column shows the segmented volume of the proposed method. These reconstructions exhibit how the proposed method was able to correctly trace thinner processes and therefore can capture the topology more effectively than the baseline algorithm. 

\subsection{Rat Visual Cortex (V1) data}
We have recently collected a relatively larger block of data of size $100 \times 100 \times 100~\mu^3$ from rat visual cortex. The images was collected at a resolution of $4 \times 4 \times 30$nm. Twp volumes of size $6 \times 6 \times 3$ and $6 \times 6 \times 4~\mu^3$ were used as training and test sets for this experiment respectively. We compared the quality of the output of the proposed method with a segmentation algorithm based on membrane prediction. Given a membrane prediction from a 3D U-net, the over-segmentation is generated in 3D by dilating the membrane probabilities to avoid potential false merges due to anisotropy. The 3D over-segmentation is then agglomerated by~\cite{parag15b}. This particular membrane predictor based segmentation method is of interest due to the the efficiency it offers us -- we have experimented rigorously to improve its accuracy.

Our experiments on the membrane probability based technique included manipulating the input size (working on downsampled version), modifying the network architecture (alternating between 2D-3D convolutions), and testing different types of filters (with or without zero-padding) to attain the desired speed and accuracy. We generated the final segmentation by applying watershed and agglomeration on the output each of these membrane predictors.

\begin{figure}[h]
\vspace{-0.1cm}
\begin{center}
\includegraphics[width=0.84\columnwidth, height=0.75\columnwidth]{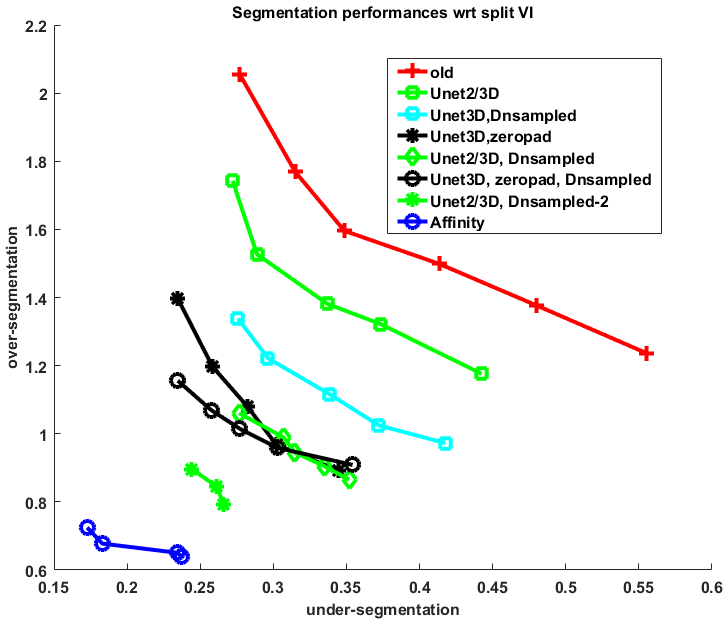}
\caption{\scriptsize Quantitative evaluation of competing methods on the $ 6 \times 6 \times 4~\mu^3 $ test volume. The x and y axes correspond to the under- and over-segmentation respectively.  Lower curves correspond to better result than the those corresponding to the curves above. The curve for the proposed method is the colored blue and is better than all other methods we tried. }\label{F:ECS_COMPARISON}
\end{center}
\vspace{-0.7cm}
\end{figure}

In Figure~\ref{F:ECS_COMPARISON}, we show the split-VI (Variation of Information) measure for each of the different predictors we tested. The x and y axes correspond to the under and over-segmentation error, as before. Each point on the curve corresponds to a threshold for the agglomeration process.  Our tests suggest that the proposed 3D affinity learning with agglomeration (blue curve) achieves the best result on this dataset. The closest technique in terms of accuracy seems to be a  Unet 2/3D with downsampled images. By 2/3D, we imply that different layers of this deep network applied the convolution with different dimensionality, i.e., some layers applies a 2D filter and some other layers the convolution is 3D. The real valued output on the downsampled images are scaled up before the subsequent operations. 

\begin{figure*}[t]
\vspace{-0.7cm}
\begin{center}
\includegraphics[width=2.2\columnwidth, height=0.52\columnwidth]{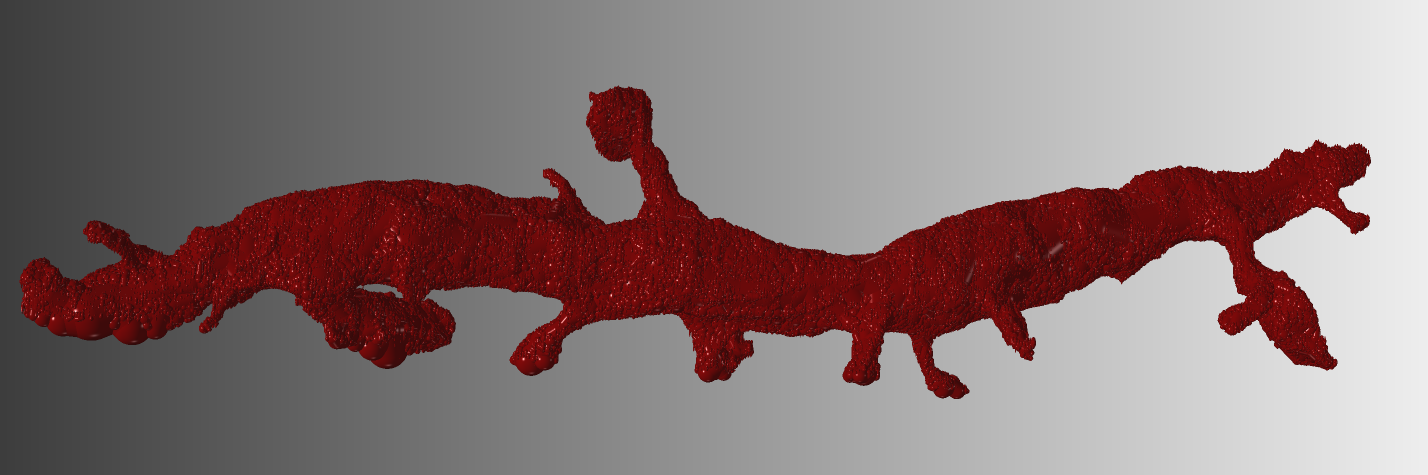}
\includegraphics[width=2.2\columnwidth, height=0.52\columnwidth]{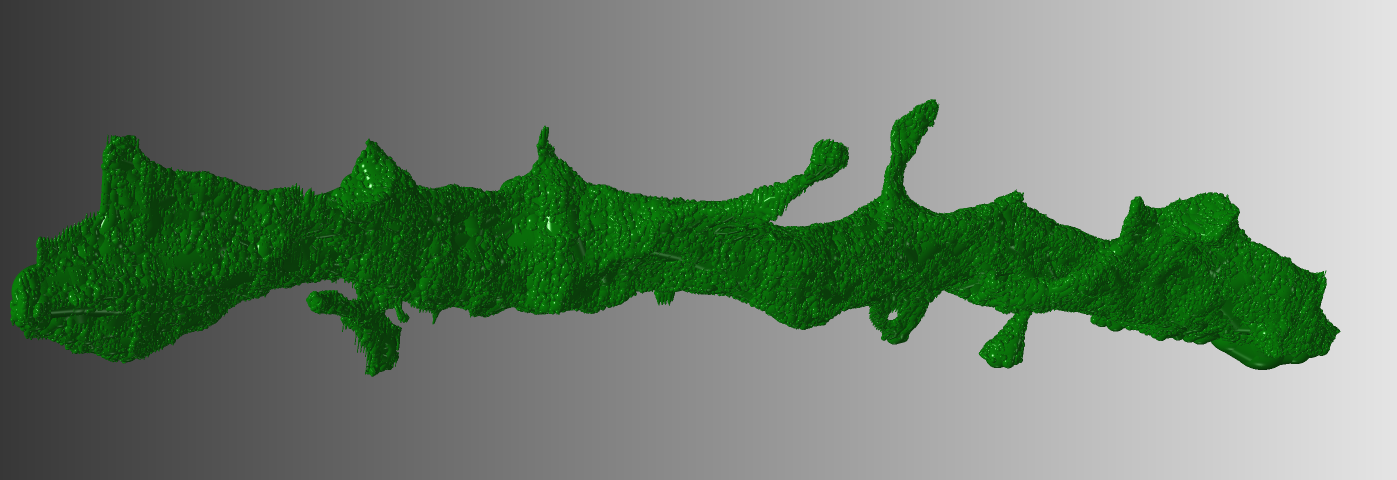}
\includegraphics[width=2.2\columnwidth, height=0.52\columnwidth]{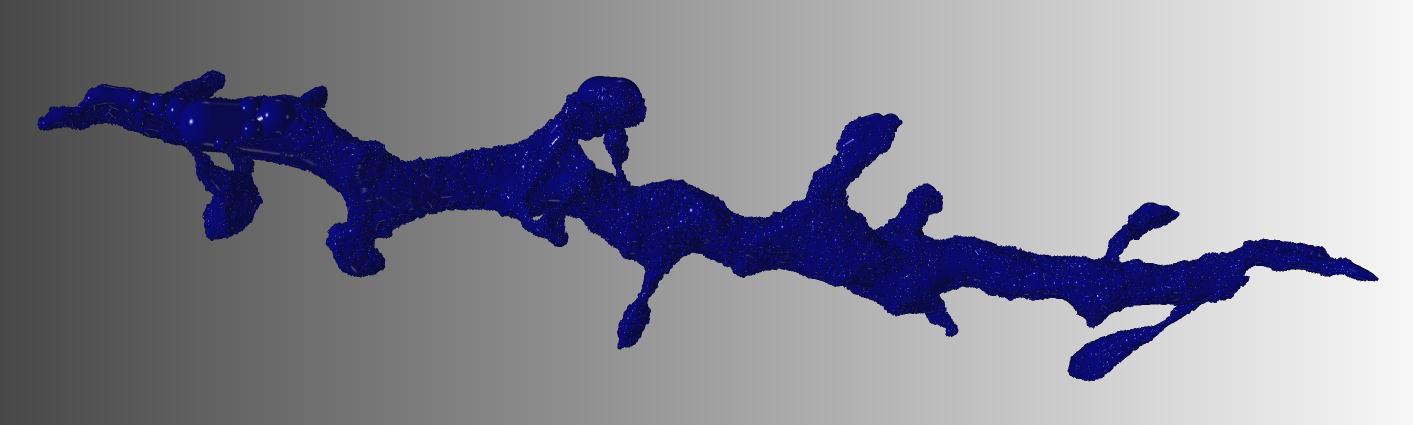}
\includegraphics[width=2.2\columnwidth, height=0.52\columnwidth]{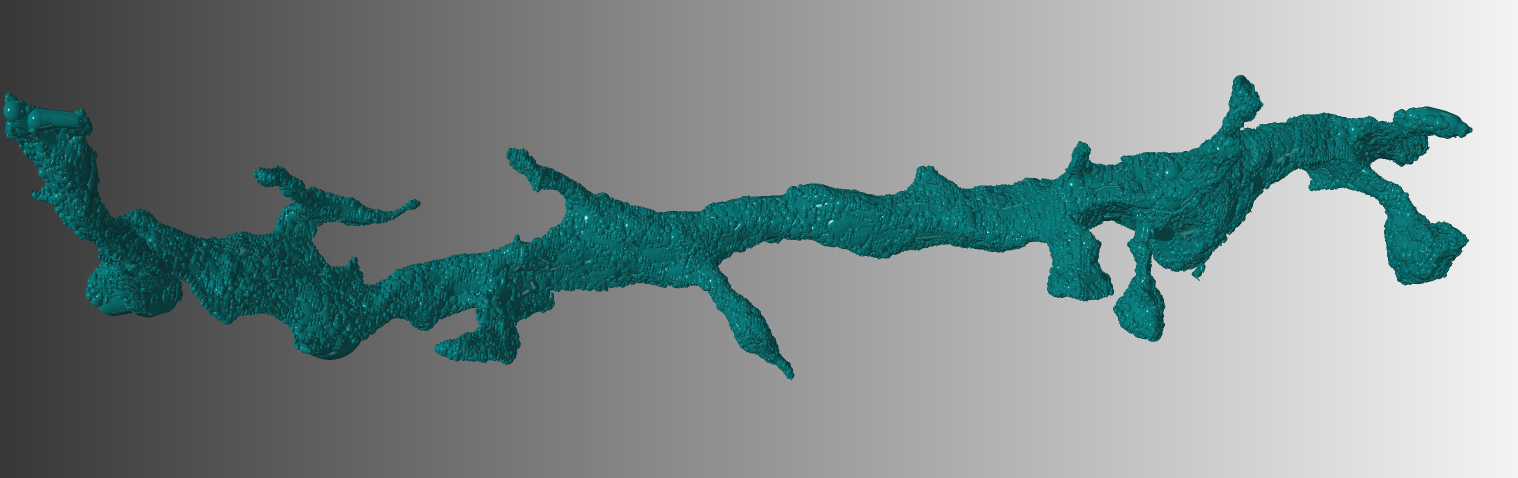}
\caption{\scriptsize 3D views of the cellular processes reconstructed by the propsed method from a $4 \times 4 \times 100 \mu^3$ rat cortex data.}\label{F:ECS_QUAL}
\end{center}
\vspace{-0.7cm}
\end{figure*}

In order to test the performance of the proposed technique on longer processes, we applied the proposed segmentation algorithm on a $4 \times 4 \times 100 \mu^3$ volume of this data. We divided this tall volume into smaller (overlapping) blocks and then stitched the segmentation across these blocks using the segmentation overlap. The 3D representation of a few largest reconstructions are demonstrated in Figure~\ref{F:ECS_QUAL}. Although the results show some false splits of small processes on these results,  we observed negligible false merges between two cellular processes. The membrane based method under-segmented more significantly compared to the proposed approach and very few of the largest reconstructions resembled a neuron part. 
\begin{figure}[h]
\vspace{-0.1cm}
\begin{center}
\includegraphics[width=0.84\columnwidth, height=0.75\columnwidth]{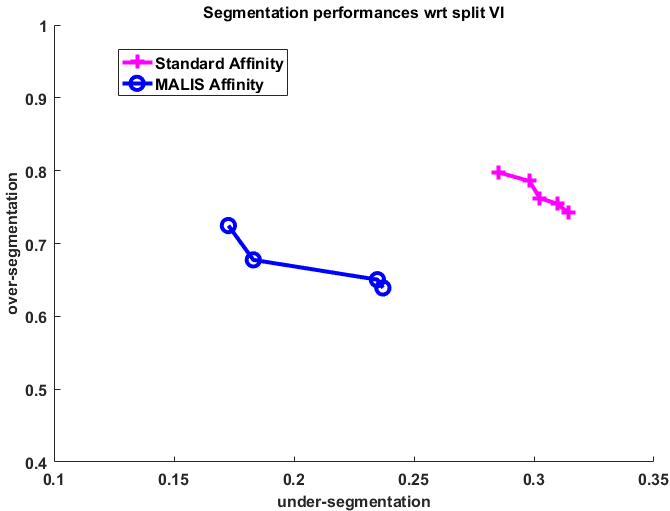}
\caption{\scriptsize Quantitative evaluation of standard and MALIS affinity learning on rat visual cortex dataset (same test $ 6 \times 6 \times 4~\mu^3 $ volume). The x and y axes correspond to the under- and over-segmentation respectively.  Lower curves correspond to better result than the those corresponding to the curves above. The MALIS learning of affinities leads to better results (blue o) than that (magenta +) produced by the affinities trained using standard method. }\label{F:ECS_MALIS_STANDARD}
\end{center}
\vspace{-0.7cm}
\end{figure}

On this dataset, we have also investigated whether or not standard affinity learning can achieve the same level of accuracy. Recall that, MALIS learning strategy identifies a sparse set of locations that are more important to maintain the neuron morphology and emphasized learning affinities at these locations. On the other hand, standard affinity learning has no such bias and tries to learn all affinities accurately. We used the same training and test volumes for this experiment. The segmentation produced by the proposed method with MALIS affinity resulted in a higher accuracy than that produced by the affinities learned in a standard fashion.
\section{Conclusion}

This paper presents an algorithm for anisotropic EM volume segmentation. Through rigorous experimentation, we have demonstrated that 3D affinities learned directly from the anisotropic images by U-net using MALIS leads to very accurate over-segmentation. This over-segmentation can then be agglomerated to produce a segmentation result that is significantly superior in quality than those generated by several existing techniques. We believe the EM connectomics community will benefit profoundly be using the proposed segmentation approach on anisotropic datasets.
{\small
\bibliographystyle{ieee}
\bibliography{affinity_agglom}
}

\end{document}